%% file: main.tex
\documentclass[10pt,twocolumn,letterpaper]{article}

\usepackage{iccv}
\usepackage{times}
\usepackage{epsfig}
\usepackage{graphicx}
\usepackage{amsmath}
\usepackage{amssymb}
\usepackage{xcolor}
\usepackage{booktabs}
\usepackage{multirow}
\usepackage{capt-of}
\usepackage{xspace}
\usepackage{colortbl}
\usepackage{xcolor}
\usepackage{textcomp}
\usepackage{algorithm}
\usepackage[accsupp]{axessibility}  %

\usepackage[breaklinks=true,bookmarks=false]{hyperref} %

\iccvfinalcopy %

\def\iccvPaperID{4018} %

\ificcvfinal\fi

\definecolor{teal}{rgb}{0.0, 0.5, 0.5}
\definecolor{goldenrod}{rgb}{0.867, 0.769, 0.255}
\definecolor{gray}{rgb}{0.843, 0.843, 0.843}
\definecolor{brown}{rgb}{0.494, 0.259, 0.0196}
\definecolor{grey}{rgb}{0.9,0.9,0.9}

\newcommand{\goldenbullet}{\textcolor{goldenrod}{\fontsize{20}{20}\selectfont\textbullet}}
\newcommand{\graybullet}{\textcolor{gray}{\fontsize{20}{20}\selectfont\textbullet}}
\newcommand{\brownbullet}{\textcolor{brown}{\fontsize{20}{20}\selectfont\textbullet}}
\newcommand{\methodname}{NeuRas\xspace}

\def \short {} %
\ifx \short \undefined

\else

\fi

\makeatletter
\g@addto@macro\@maketitle{
\vspace{-2.5em}
\begin{figure}[H]
\setlength{\linewidth}{\textwidth}
\setlength{\hsize}{\textwidth}
\centering
\includegraphics[trim={0cm, 0cm, 0cm, 0.0cm},clip,width=\linewidth]{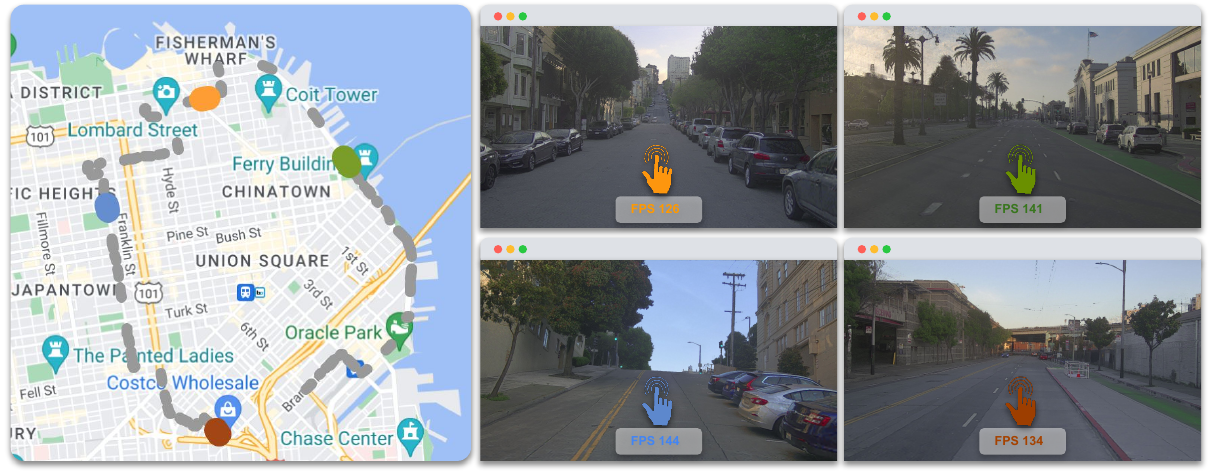}
\vspace{-4.5mm}
\captionof{figure}{
\textbf{Neural Scene Rasterization.}
Our method renders urban driving scenes \textbf{(1920$\times$1080)} at high quality and \textbf{$>$100 FPS} by leveraging neural textures and fast rasterization.
We reconstruct driving scenes in Bay Area and show the rendering at four streets on the map.
Please refer to our project page \url{https://waabi.ai/neuras/} for more results on driving scenes as well as drone scenes.
}
\vspace{1mm}
\label{fig:teaser}
\end{figure}
}
\makeatother

\begin{document}

\title{Real-Time Neural Rasterization for Large Scenes}

\author{
Jeffrey Yunfan Liu$^{1, 3\dag}$
\quad Yun Chen$^{1, 2}$\thanks{Indicates equal contribution. \textsuperscript{\dag}Work done while an intern at Waabi.}
\quad Ze Yang$^{1, 2*}$
\quad Jingkang Wang$^{1, 2}$
\quad Sivabalan Manivasagam$^{1, 2}$ \\
\quad Raquel Urtasun$^{1, 2}$ \\
\normalsize{
$^{1}$Waabi
\quad $^{2}$University of Toronto
\quad $^{3}$University of Waterloo} \\
{\small\texttt{\{jliu,  ychen, zyang, jwang, siva, urtasun\}@waabi.ai}}
}

\maketitle
\ificcvfinal\thispagestyle{empty}\fi

\begin{abstract}
We propose a new method for realistic real-time novel-view synthesis (NVS) of large scenes.
Existing
neural rendering methods generate realistic results, but primarily work for small scale scenes ($<50m^2$) and have difficulty at large scale ($>10000m^2$).
Traditional graphics-based rasterization rendering is fast for large scenes but lacks realism and requires expensive manually created assets.
Our approach combines the best of both worlds by taking a moderate-quality scaffold mesh as input and learning a neural texture field and shader to model view-dependant effects to enhance realism, while still using the standard graphics pipeline for real-time rendering.
Our method outperforms existing neural rendering methods, providing at least 30$\times$ faster rendering with comparable or better realism for large self-driving and drone scenes.
Our work is the first to enable real-time rendering of large real-world scenes.
\end{abstract}

\input{tex/1-intro.tex}

\input{tex/2-related_work.tex}

\input{tex/3-method.tex}

\input{tex/4-experiment.tex}

\input{tex/5-conclusion.tex}

{\small
\bibliographystyle{ieee_fullname}
\bibliography{ref}
}

\end{document}

%% file: tex/1-intro.tex
\section{Introduction}
Synthesizing and rendering images for large-scale scenes, such as city blocks,
holds significant
value in
fields such as robotics simulation and virtual reality (VR).
In these fields, achieving a high level of
realism and speed is of utmost importance.
VR requires photorealistic renderings at interactive frame rates for an immersive and seamless user experience.
Similarly, robot simulation development requires high-fidelity image quality for real world transfer and high frame rates for evaluation and training at scale, especially for closed-loop
sensor
simulation~\cite{szot2021habitat}.

Achieving both speed and realism in large-scale scene synthesis has been a long-standing challenge.
Recently, a variety of neural rendering approaches~\cite{mildenhall2020nerf,riegler2020fvs,riegler2021svs} have shown impressive realism results in novel view synthesis (NVS).
These methods fall into two primary paradigms:
implicit and explicit-based approaches.
Implicit-based approaches~\cite{mildenhall2020nerf,barron2022mip,niemeyer2020differentiable,yang2023reconstructing} represent scene geometry and appearance with multi-layer-perceptrons (MLPs) and render novel views by evaluating these MLPs hundreds of thousands of times via volume rendering.
Explicit-based approaches~\cite{riegler2020fvs,riegler2021svs} reconstruct a geometry scaffold (e.g., mesh, point cloud), and then learn image features that are blended and refined with neural networks (NN) to render a novel view.
Both implicit and explicit methods require large amounts of NN computation to perform NVS.
As a consequence, these approaches have primarily focused on reconstructing objects or small-scale scenes ($<50m^2$), and typically render at non-interactive frame rates ($< 5$
FPS).

Several recent methods have enabled rendering at higher frame rates ($> 20$ FPS) while maintaining realism through several strategies such as sub-dividing the scene~\cite{mueller2022instant, yu2021plenoctrees, sun2021direct,turki2022mega,reiser2021kilonerf}, caching mechanisms~\cite{garbin2021fastnerf, hedman2021baking, yu2021plenoctrees,hu2022efficientnerf}, and optimized sampling~\cite{wu2022diver, neff2021donerf, kondo2021vaxnerf}.
However, these approaches still focus primarily on single objects or small scenes.
They either do not work on large scenes ($>10000 m^2$) or far-away regions due to memory limitations when learning and rendering such a large volume, or have difficulty maintaining both photorealism and high speed.
One area where large scenes are rendered at high speeds is in computer graphics through rasterization-based game engines~\cite{karis2013real}.
However, to render with high quality, these engines typically require accurate specifications of the exact geometry, lighting, and material properties of the scene, along with well-crafted shaders to model physics.
This comes at the cost of tedious and time-consuming manual efforts from expert 3D artists and expensive and complex data collection setups for dense capture~\cite{egels2001digital, guo2019relightables}.
While several recent methods have leveraged rasterization-based rendering in NVS~\cite{cao2022fwd,chen2022mobilenerf,aliev2020neural,rakhimov2022npbg++,lombardi2021mixture,kellnhofer2021neural}, they have only been demonstrated on small scenes. %

In this paper, we introduce \methodname, a novel neural
rasterization
approach that combines rasterization-based graphics and neural texture representations for realistic real-time rendering of large-scale scenes.
Given a sequence of sensor data (images and optionally LiDAR), our key idea is to first build a moderate quality geometry mesh of the scene, easily generated with existing 3D reconstruction methods~\cite{unisim,schonberger2016structure,Yu2022MonoSDF,mueller2022instant}.
Subsequently, we perform rasterization with learned feature maps and Multi-Layer Perceptrons (MLPs) shaders to model view-dependent effects.
Compared to computationally expensive neural volume rendering, leveraging an approximate mesh enables high-speed rasterization, which scales
well for large scenes.
Compared to existing explicit-based geometry methods that use large neural networks to perform blending and image feature refinement, we use light-weight MLPs that can be directly exported as fragment shaders in OpenGL for real-time rendering.
We also design our neural rasterization method with several enhancements to better handle large scenes.
First, inspired by multi-plane and multi-sphere image representations~\cite{zhou2018stereo, attal2020matryodshka}, we model far-away regions with multiple neural skyboxes to enable rendering of
distant buildings and sky.
Additionally, most NVS methods focus on rendering at target views that are close to the source training views.
But for simulation or VR, we need NVS to generalize to novel viewpoints that deviate from the source views.
To ensure our approach works well at novel viewpoints, we utilize vector quantization~\cite{he2022masked, van2017neural} to make neural texture maps more robust and to store them efficiently.

Experiments on large-scale self-driving scenes and drone footage demonstrate that \methodname achieves the best trade-off between speed and realism
compared to existing SoTA.
Notably,
\methodname can achieve comparable performance to %
NeRF-based methods while being at least
$30 \times$
faster.
To the best of our knowledge, \methodname is the first method of its kind that is capable of realistically rendering large scenes at a resolution of
$1920 \times 1080$
in real-time, enabling more scalable and realistic rendering for robotics and VR.

%% file: tex/2-related_work.tex
\section{Related Work}

\begin{figure*}[]
    \centering
     \includegraphics[width=1.0\linewidth]{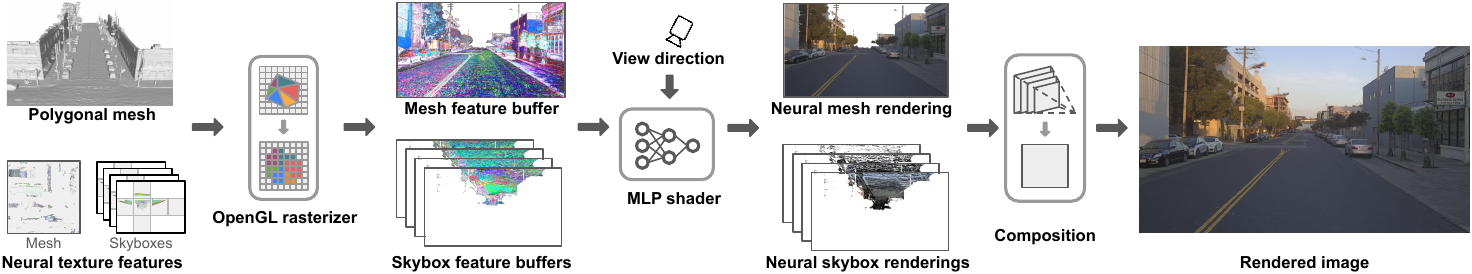}
     \caption{\textbf{\methodname pipeline.}
     We first rasterize screen-space feature buffers from the scene representation. Then a learned MLP shader takes the rasterized feature buffers and view-direction as inputs and predicts a set of rendering layers. Finally the rendered layers are composited to synthesize the RGB image.
     }
     \label{fig:rendering}
\end{figure*}

\paragraph{Explicit-based synthesis:}
Classical 3D reconstruction methods such as structure-from-motion and multi-view stereo~\cite{schoenberger2016sfm, schoenberger2016mvs,agarwal2011building} have been extensively utilized to reconstruct large-scale 3D models of real-world scenes.
However, texture mapping applied during the reconstruction process often produces unsatisfactory appearance due to resolution limitations and lack of view-dependence modelling.
To address the 
challenges
of NVS, image-based rendering methods employ the reconstructed geometry as a proxy to re-project target views onto source views, and then blend the source views heuristically or by using a convolutional network~\cite{buehler2001unstructured, riegler2020fvs,chen2021geosim,cao2022fwd,wang2022cadsim}.  
These methods typically require a large amount of memory and may still have visual aberrations due to errors in image blending.
Alternatively, point-based methods~\cite{li2022read,aliev2020neural,rakhimov2022npbg++} use per-point feature descriptors and apply convolutional networks to produce images. 
However, these methods are inefficient in rendering due to large post-processing networks and often exhibit visual artifacts when the camera moves.
Similarly, multi-plane images~\cite{choi2018extreme,zhou2018stereo,flynn2019deepview,mildenhall2019local} or multi-sphere images~\cite{attal2020matryodshka} are applied for outdoor-scenes rendering in real-time, but they can only be rendered with restricted camera motion.
Our work builds on these techniques by leveraging an explicit geometry, but then utilizes UV neural texture maps and fast rasterization to boost speed and realism.

\paragraph{Implicit-based synthesis:}
In recent years, implict neural field methods, especially NeRF~\cite{mildenhall2020nerf}, have become the foundation for many state-of-the-art NVS techniques. NeRF represents the scene 
as an MLP
that is
optimized based on camera observations. To accelerate reconstruction time, several methods have been proposed~\cite{mueller2022instant,yu2021plenoxels,sun2021direct,chen2022tensorf,lin2022enerf}, but cannot achieve real-time rendering for high resolution.
To accelerate rendering speed, ``baking" methods~\cite{reiser2021kilonerf,garbin2021fastnerf,yu2021plenoctrees,hedman2021baking,wu2022diver} typically pre-compute and store the neural field properties for rendering.
However, these methods require a large amount of memory and disk storage, limiting their applicability mainly to small objects.
Several recent works share similar ideas with our approach of utilizing the graphics pipeline to accelerate neural rendering while maintaining realism.
MobileNeRF~\cite{chen2022mobilenerf} also uses an explicit mesh with UV feature representations and an MLP baked in GLSL. However, its grid-mesh representation limits its scalability to large scenes due to memory constraints, and its hand-crafted mesh configurations fail to adapt to complex outdoor scenes.
Similarly, BakedSDF~\cite{yariv2023bakedsdf} requires a high-resolution mesh that occupies a significant amount of space even for smaller scenes, and MeRF~\cite{reiser2023merf} bakes NeRF to tri-plane images, which reduces memory consumption but limits its scalability to large scenes.
In contrast, our approach only requires a moderate-quality mesh as a proxy and uses a quantized UV texture feature, which can be easily extended to large scenes to achieve high realism in real-time.

\paragraph{Large-scene synthesis:}
Several methods have been proposed to extend NeRF for
large-scene synthesis.
BlockNeRF~\cite{Tancik_2022_CVPR_blocknerf} and MegaNeRF~\cite{turki2022mega} divide large scenes into multiple spatial cells to increase the model capacity. 
Recent work GP-NeRF~\cite{zhang2023efficient} further improves the training speed by using hybrid representations.
READ~\cite{li2022read} uses LiDAR point cloud descriptors followed by a convolutional network to synthesize autonomous scenes. 
BungeeNeRF~\cite{xiangli2022bungeenerf} employs a learnable progressive growing model for city-scale scenes. 
These methods typically struggle with high computational costs, especially for real-time rendering.
For interactive visualization, URF~\cite{rematas2021urban} bakes mesh while MegaNeRF applies techniques like caching and efficient sampling, but the realism drops significantly.
To the best of our knowledge, our approach is the first to demonstrate realistic rendering in real-time for such large-scale scenes.

%% file: tex/3-method.tex
\section{Neural Scene Rasterization}
In this section, we describe \methodname, which aims to perform real-time rendering of large-scale scenes.
Given a set of posed images and a moderate-quality reconstructed mesh, our method generates a scene mesh with neural texture maps and view-dependent fragment shaders.
Using the initial mesh, we first generate a UV parameterization to learn neural textures on.
We then jointly learn a discrete texture feature codebook and view-dependent lightweight MLPs that can effectively represent scene appearance.
Finally, we bake the texture feature codebook and the MLPs into a set of neural texture maps and a fragment shader that can be run in real time with existing graphics pipelines.
We now first introduce our approach for representing large-scale scenes (Sec.~\ref{sec:scene}), then describe how we render and learn the scene (Sec.~\ref{sec:rendering}-\ref{sec:learning}), and finally how we export our model into real time graphics pipelines (Sec.~\ref{sec:realtime}).

\subsection{Scene Representation}
\label{sec:scene}
In this paper 
our focus is on
rendering large-scale outdoor scenes. In order to handle potentially infinite depth ranges (\eg, sky, vegetation, mountain, \etc) as well as nearby regions, we utilize a hybrid approach.
The entire 3D scene is partitioned into two regions: an inner cuboid region (\textit{foreground}) modelled by a polygonal mesh textured with neural features, and an outer cuboid region (\textit{background}) modelled by neural skyboxes.
Such a hybrid scene representation allows us to model fine-grained details in both close-by regions and far-away regions, and enables rendering with a remarkable degree of camera movement.

\paragraph{Foreground representation:}
For the 
\textit{foreground}
region, we leverage an explicit geometry mesh scaffold to learn and render neural textures.
Our approach can be applied on various sources of mesh, for example we can leverage existing neural reconstruction methods~\cite{unisim,mueller2022instant} or SfM~\cite{schoenberger2016sfm} (see supp. materials for details). %
Initially, the reconstructed mesh could contain over tens of millions of triangle faces, which represents the geometry well, but may have self-intersections and duplicate vertices.
We thus preprocess it to reduce computational cost and improve UV mapping quality.
We first cluster nearby vertices together and perform quadric mesh decimation~\cite{garland1997surface} to simplify the mesh while 
preserving
essential structure, and then perform face culling to remove non-visible triangle faces (\wrt source camera views).
Finally we utilize a UV map generation tool~\cite{xatlas} to unfold the mesh to obtain the UV mappings for each of its vertices.
The resulting triangle mesh $\mathcal M = \{\mathbf v, \mathbf t, \mathbf f\}$ consists of vertex positions $\mathbf v \in \mathbb R^{N\times 3}$, vertex UV coordinates $\mathbf t \in \mathbb R^{N\times 2}$, and a set of triangle faces $\mathbf f$.
Based on the generated UV mapping, we initialize a learnable UV feature map $\mathbf T \in \mathbb R^{V \times U \times D}$ to represent the scene appearance covered by the mesh.
Using neural features instead of a color texture map enables modelling view-dependent effects during rendering (see Sec. \ref{sec:rendering}).

\paragraph{Background representation:}
It is challenging to model the far-away 
\textit{background}
regions with polygonal mesh because of the complexity and scale of that region.

As an alternative, we draw inspiration from the concept of multi-plane images~\cite{zhou2018stereo,flynn2019deepview,mildenhall2019local} and multi-sphere images~\cite{attal2020matryodshka} to represent the \textit{background} region using neural skyboxes.

The neural skyboxes represents the scene as a set of cuboid layers, and each layer $\mathcal S_i = \{\mathbf S_i^1, \cdots, \mathbf S_i^6\}$ contains 6 individual feature maps that each 
represents
one plane of the cuboid.
The feature maps represent both geometry and view-dependent appearance 
of the scene. 
Our neural skyboxes can represent a wide range of depths and 
can
be integrated in existing graphics pipelines~\cite{karis2013real}, enabling efficient rendering.

\subsection{Rendering Large Scenes}
\label{sec:rendering}
Fig.~\ref{fig:rendering} shows an overview of our rendering pipeline. Our \methodname framework is inspired by the deferred shading pipeline from real-time graphics~\cite{thies2019deferred}.
We first rasterize the foreground mesh and neural skyboxes with neural texture maps to the desired view point, producing a set of image feature buffers.
The feature buffers are then processed with MLPs to produce a set of rendering layers, which are composited to synthesize the final RGB image.
We now describe this 
rendering
process in more detail.

\paragraph{Foreground rendering:}
Given the camera pose and intrinsics, we first rasterize the mesh into screen space, obtaining a UV coordinate $(u, v)$ for each pixel $(x, y)$ 
on the screen.
We then sample the UV feature map $\mathbf T \in \mathbb R^{V \times U \times D}$ using the rasterized UV coordinates and obtain a feature buffer $\mathbf F_0 \in \mathbb R^{H \times W \times D}$:
\begin{align}
	\mathbf F_0(x, y) = \texttt{BilinearSample}(u, v, \mathbf T),
\end{align}
where $V \times U$ is the UV feature map resolution, $H \times W$ is the rendering resolution, and $D$ is the feature dimension.
In addition to a feature buffer, the rasterization also generates a opacity mask $\mathbf O_0 \in \mathbb R^{H \times W}$ to indicate if a pixel is covered by the polygonal mesh.
To render the RGB image, we concatenate the rendered feature with the view direction $\mathbf d(x, y)$ and pass through a learnable MLPs shader $f_{\theta^{\mathbf T}}$:
\begin{align}
	\mathbf I_0(x, y) = f_{\theta^{\mathbf T}}\left(\mathbf F_0(x, y), \mathbf d(x, y)\right),
	\label{eqn:mesh_render}
\end{align}
where $\theta^{\mathbf T}$ is the MLP parameters, $\mathbf I_0(x, y)$ is the rendered RGB color for pixel $(x, y)$.

\paragraph{Background rendering:}
To render the neural skybox feature layers $\{\mathcal S_i\}_{i=1}^L$ representing distant \textit{background} regions, we project camera ray shooting each pixel $(x, y)$ and compute its intersection points $\{\mathbf p_i\}_{i=1}^L$ with layers $1$ to $L$, from near to far.
Next, we sample the features $\{\mathbf f_i\}_{i=1}^L$ corresponding to the intersection points on the neural skybox feature map at each layer, generating a set of feature buffers $\mathbf F_i \in \mathbb R^{H \times W \times D}$, where $i=1, \cdots, L$.
This step can be efficiently performed with the OpenGL rasterizer.
We then use a learnable MLPs shader $f_{\theta^{\mathbf S}}$ to process the feature buffers and outputs the opacity map $\mathbf O_i \in \mathbb R^{H \times W}$ and RGB color map $\mathbf I_i \in \mathbb R^{H \times W \times 3}$ for each layer:
\begin{align}
	\mathbf O_i(x, y), \mathbf I_i(x, y) = f_{\theta^{\mathbf S}}\left(\mathbf F_i(x, y), \mathbf d(x, y) \right),
\end{align}
where $\theta^{\mathbf S}$ represents the parameters of the MLPs.
The MLPs shader first processes the input feature $\mathbf F_i(x, y)$, and outputs opacity $\mathbf O_i(x, y)$ and an intermediate feature vector.
The feature vector is then concatenated with $\mathbf d(x, y)$, the view direction of camera ray,  and passed to the last layers that output the view-dependent color $\mathbf I_i(x, y)$.

\paragraph{Compositing foreground and background:}
To synthesize the final RGB image, we composite the rendered layers from the foreground mesh $\{ \mathbf I_0(x, y), \mathbf O_0(x, y) \}$ and neural skyboxes $\{ \mathbf I_i(x, y), \mathbf O_i(x, y) \}_{i=1}^L$ by repeatedly compositing the RGB and opacity layers, from near to far:
\begin{align}
	\mathbf I(x, y) = \sum_{i=0}^L \mathbf I_i(x, y) \cdot \mathbf O_i(x, y) \cdot \prod_{j=0}^{i-1} (1-\mathbf O_j(x, y)).
	\label{eqn:rgb_render}
\end{align}
The term $\mathbf{I}_i(x, y) \cdot \mathbf{O}_i(x, y)$ represents the color contribution of the current layer $i$, and
$\prod_{j=1}^{i-1} (1-\mathbf O_j(x, y))$ denotes the fraction of the color that will remain after attenuation through the layers in front.
This compositing process ensures that the RGB values are correctly blended.

\begin{figure}[t]
    \centering
     \includegraphics[width=1.0\linewidth]{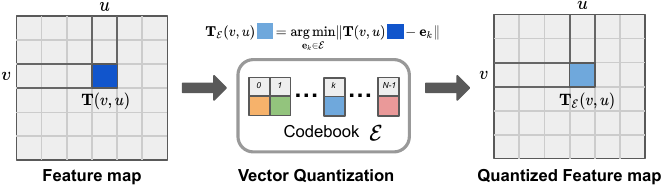}
     \caption{\textbf{Quantized feature representation.}
     For each entry in the feature map, we map it to the closest learnable latent code $\mathbf e_k$ in the codebook $\mathcal E$.
     }
     \label{fig:quantization}
\end{figure}

\paragraph{Quantized texture representation:}
To 
encourage
the sharing of latent features in visually similar regions such as roads and sky, we apply vector quantization (VQ) to regularize the neural texture maps.
This allows these features to be supervised from a large range of view directions which improves view-point extrapolation performance. Furthermore, it also significantly compacts the feature representations, reducing the offline storage space.
We follow~\cite{van2017neural} and maintain two codebooks $\mathcal E^{\mathbf T}$ and $\mathcal E^{\mathbf S}$, each consists of $K$ learnable latent code $\mathbf e_k \in \mathbb R^D$, with $k=1, \cdots, K$.
In the forward pass, we quantize the UV feature map $\mathbf{T}_{\mathcal E} \in \mathbb R^{V \times U \times D}$ and neural skyboxes feature maps $\mathbf{S}_{\mathcal E} \in \mathbb R^{L \times V \times U \times D}$ by mapping each feature to its closest latent code in the codebook:
\begin{equation}
\begin{aligned}
	\mathbf T_{\mathcal E} (v, u) &= \underset{\mathbf e_k \in \mathcal E^{\mathbf T}}{\arg\min} \lVert \mathbf{T} (v, u) - \mathbf e_k \rVert, \\
	\mathbf S_{\mathcal E} (l, v, u) &= \underset{\mathbf e_k \in \mathcal E^{\mathbf S}}{\arg\min} \lVert \mathbf{S} (l, v, u) - \mathbf e_k \rVert,
\end{aligned}
\label{eqn:quantization}
\end{equation}
where $v, u$ are the spatial coordinates of the feature map, and $l$ is the layer index of the neural skyboxes.
Fig.~\ref{fig:quantization} shows the feature map quantization process.
We use the quantized features to compute the synthesized image in Eqn.~\ref{eqn:rgb_render}.

\subsection{Learning \methodname}
\label{sec:learning}
We jointly optimize the feature map $\mathbf T$, $\mathbf S$, the codebook $\mathcal E^{\mathbf T}, \mathcal E^{\mathbf S}$, as well as the parameters $\theta^{\mathbf T}, \theta^{\mathbf S}$ of the MLP shaders by minimizing the photometric loss and perceptual loss between our rendered images and camera observations, as well as the VQ regularizer.
Our full objective is:
\begin{align}
		\mathcal L =\mathcal L_\text{rgb} + \lambda_\text{perc}\mathcal L_\text{perc} + \lambda_\text{vq}\mathcal L_\text{vq}.
\end{align}
In the following, we discuss each loss term in more detail.

\paragraph{Photometric loss:}
$\mathcal L_\text{rgb}$ measures the $\ell_2$ distance between the rendered and the observed images, the loss is defined as:
\begin{align}
	\mathcal L_\text{rgb} = \lVert \mathbf I - \mathbf{\hat I} \rVert_2,
\end{align}
where $\mathbf I$ is the rendered image from Eqn.~\ref{eqn:rgb_render} and $\mathbf{\hat I}$ is the corresponding observed camera image.

\paragraph{Perceptual loss:}
We use an additional perceptual loss~\cite{zhang2018unreasonable,wang2018pix2pixHD} to enhance the rendered image quality.
This loss
measures the ``perceptual similarity" that is more consistent with human visual perception:
\begin{align}
	\mathcal L_\text{perc} = \sum_{i=1}^M \frac{1}{N_i} \left\lVert V_i(\mathbf I) - V_i(\mathbf{\hat I}) \right\rVert_1,
\end{align}
where $V_i$ denotes the i-th layer with $N_i$ elements of the pre-trained VGG Network~\cite{simonyan2014very}.

\paragraph{VQ loss:}
To update the codebook $\mathcal E$, we follow~\cite{van2017neural} and define the 
VQ
loss term as:
\begin{equation}
\begin{aligned}
	\mathcal L_\text{vq}
	&=
	\lVert\operatorname{sg}[\mathbf T] - \mathbf T_{\mathcal E}\rVert_2^2 +
	 \lVert\operatorname{sg}[\mathbf S] - \mathbf S_{\mathcal E}\rVert_2^2 \\
	&+
	\beta \lVert\operatorname{sg}[\mathbf T_{\mathcal E}] - \mathbf T\rVert_2^2 +
	\beta \lVert\operatorname{sg}[\mathbf S_{\mathcal E}] - \mathbf S\rVert_2^2,
\end{aligned}
\end{equation}
where $\operatorname{sg}[\cdot]$ denotes the stop-gradient operator that behaves as the identity map at forward pass and has zero partial derivatives at backward pass.
The first two terms form the alignment loss and encourage the codebook latents to follow the feature maps. The 
last
two terms form the commitment loss which stabilizes training by discouraging the features from learning much faster than the codebook.
It is noted that the quantization step in Eqn.~\ref{eqn:quantization} is non-differentiable.
We approximate the gradient of the feature maps $\mathbf T, \mathbf S$ using the straight-through estimator~\cite{bengio2013estimating}, which simply passes the gradient from the quantized feature to the original feature unaltered during back-propagation.

\subsection{Real Time Rendering}
\label{sec:realtime}
To enable \methodname to render in real time, we convert our scene representations and MLPs to be compatible with the graphics rendering pipeline.
The mesh, skyboxes, and texture representations are all directly compatible with the OpenGL, while the learned MLPs $f_{\theta^{\mathbf T}}$ and $f_{\theta^{\mathbf S}}$ are converted to fragment shaders in OpenGL. 
During each rendering pass, the triangle mesh $\mathcal M$ and the skyboxes $\{\mathcal S_i\}_{i=1}^L$ are rasterized to the screen as a set of fragments, and each fragment is associated with a feature vector that is  bilinearly sampled from the neural texture maps. 
The fragment shader then maps each fragment's features to RGB color and opacity. To ensure correct alpha compositing, the scene mesh and cuboids are sorted depth-wise and rendered from back to front, following the procedure outlined in Eqn.~\ref{eqn:rgb_render}.

%% file: tex/4-experiment.tex
\begin{figure}[]
	\centering
	{\includegraphics[width=0.8\linewidth]{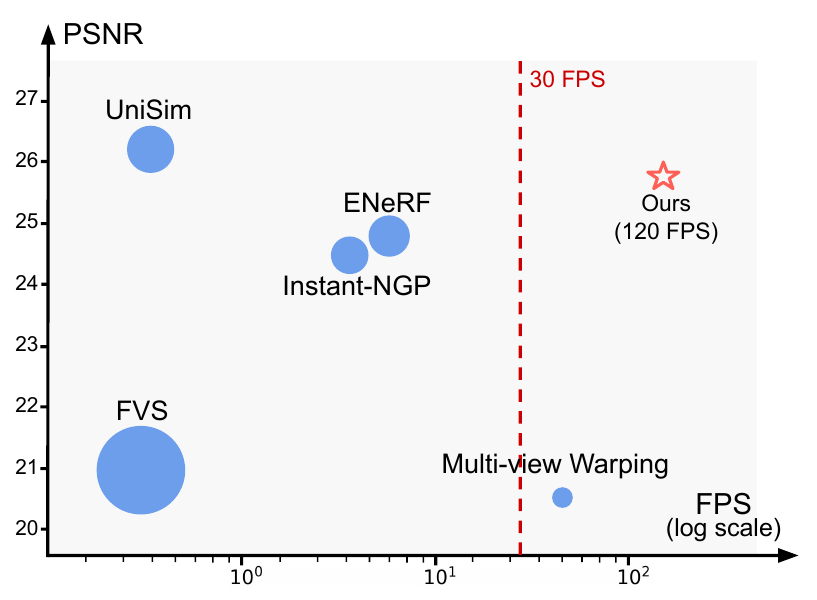}}
	\caption{\textbf{Rendering realism vs efficiency.} Our method achieves the best tradeoff between realism and speed. The size of the markers indicates the memory consumption required for rendering.}
	\vspace{-5pt}
	\label{fig:tradeoff}
  \end{figure}

\begin{figure*}[t]
    \centering
     \includegraphics[width=1.0\linewidth]{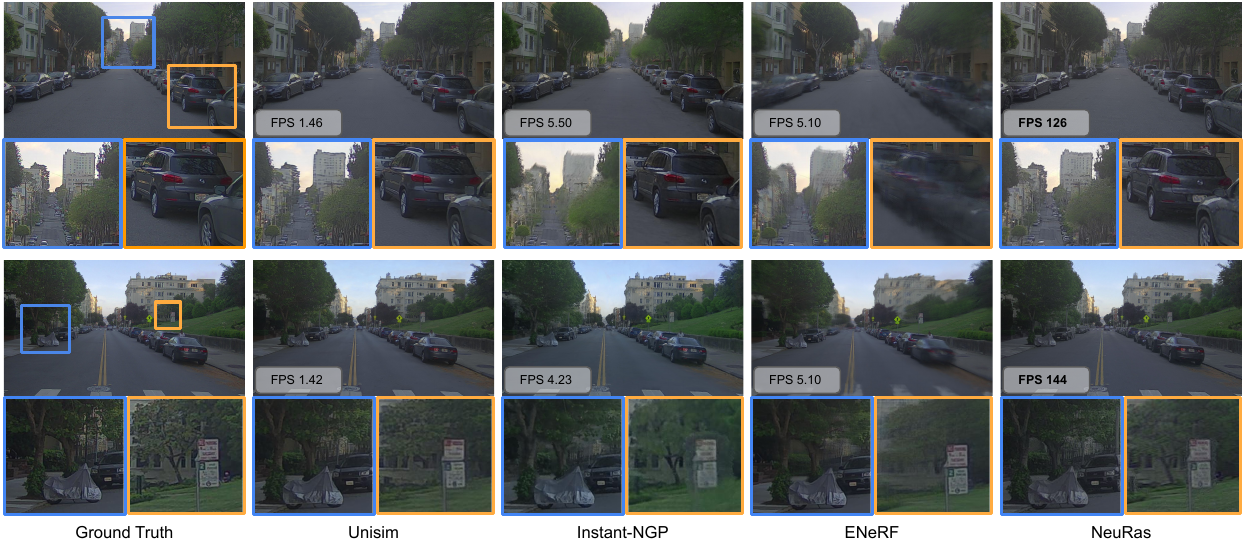}
     \caption{
     \textbf{Qualitative results on driving scenes.} Compared to existing
     novel view synthesis approaches, \methodname produces competitive realism and achieves real-time rendering ($> 100$ FPS).
     }
     \label{fig:qual_panda}
\end{figure*}

\begin{table*}[ht]
	\centering
	\resizebox{0.99\textwidth}{!}{
		\begin{tabular}{llllllllll}
			\toprule
			\multirow{2}{*}{Methods} & \multicolumn{4}{c}{Interpolation} &
			\multicolumn{2}{c}{Lane Shift} & \multicolumn{2}{c}{Resources} \\
			\cmidrule(r){2-5} \cmidrule(l){6-7} \cmidrule(l){8-9} & {MSE$\downarrow$ }
			& {PSNR$\uparrow$ } & {SSIM$\uparrow$ } & {LPIPS$\downarrow$ } & {FID$\downarrow$ @2m} & {FID$\downarrow$ @3m} & Memory (MB) $\downarrow$ & FPS$\uparrow$  \\
			\midrule
			Instant-NGP~\cite{mueller2022instant} & 0.0035 & 24.76 & 0.79 \raisebox{-0.6ex}{\goldenbullet} & 0.40 & 134.20&135.75 &  7491 &3.2 \\
			ENeRF~\cite{lin2022enerf} &0.0034 \raisebox{-0.7ex}{\brownbullet} & 24.92 \raisebox{-0.7ex}{\brownbullet} & 0.77 & 0.34 & 190.96 & 243.93 & 7285 \raisebox{-0.7ex}{\brownbullet} & 5.1 \raisebox{-0.7ex}{\brownbullet}\\
			UniSim~\cite{unisim} & 0.0030 \raisebox{-0.6ex}{\goldenbullet} & 26.28 \raisebox{-0.6ex}{\goldenbullet}  & 0.78 \raisebox{-0.6ex}{\graybullet} & 0.26 \raisebox{-0.6ex}{\goldenbullet}  & 92.78 \raisebox{-0.6ex}{\goldenbullet}  & 100.63 \raisebox{-0.6ex}{\goldenbullet}  & 7623 & 1.4 \\ \midrule

			Multi-View Warping~\cite{chaurasia2013depth} & 0.0095 & 20.55 & 0.64 & 0.39 & 164.09 & 177.19 & 1441 \raisebox{-0.7ex}{\goldenbullet} & 50 \raisebox{-0.7ex}{\graybullet} \\
			FVS~\cite{riegler2020fvs} & 0.0089 & 20.87 & 0.71 & 0.29   \raisebox{-0.7ex}{\graybullet}  & 116.36 \raisebox{-0.7ex}{\brownbullet} & 122.87 \raisebox{-0.7ex}{\brownbullet} & 15243 & 0.3 \\ \midrule
			 Ours &0.0030 \raisebox{-0.6ex}{\goldenbullet} & 25.45 \raisebox{-0.7ex}{\graybullet}  &0.75 \raisebox{-0.7ex}{\brownbullet} & 0.31 \raisebox{-0.7ex}{\brownbullet} & 105.64  \raisebox{-0.7ex}{\graybullet} & 111.63  \raisebox{-0.7ex}{\graybullet} & 4538 \raisebox{-0.7ex}{\graybullet} & 120 \raisebox{-0.7ex}{\goldenbullet} \\
			\bottomrule
	\end{tabular}}
	\vspace{5pt}
	\caption{
	\textbf{State-of-the-art comparison on PandaSet.} Our method synthesizes novel views  (1920$\times$1080) in real time with high visual quality on urban driving scenes.
	We mark the methods with best performances using gold  \raisebox{-0.7ex}{\goldenbullet}, silver \raisebox{-0.7ex}{\graybullet}, and bronze \raisebox{-0.7ex}{\brownbullet} medals.
	}
	\label{tab:sota_panda}
\end{table*}

\section{Experiments}
In this section, we introduce our experimental setting, and then compare our approach with state-of-the-art NVS methods on large-scale driving scenes and drone footages.
We demonstrate that our neural rendering system achieves the best balance between photorealism and rendering efficiency.
We then ablate our design choices, showing the value of the neural shader and vector quantization for improved realism and extrapolation robustness.
Finally, we show \methodname can speed up various NeRF approaches in a plug-and-play fashion by leveraging their extracted meshes for real-time NVS of large scenes.

\begin{figure*}[ht]
    \centering
     \includegraphics[width=1.0\linewidth]{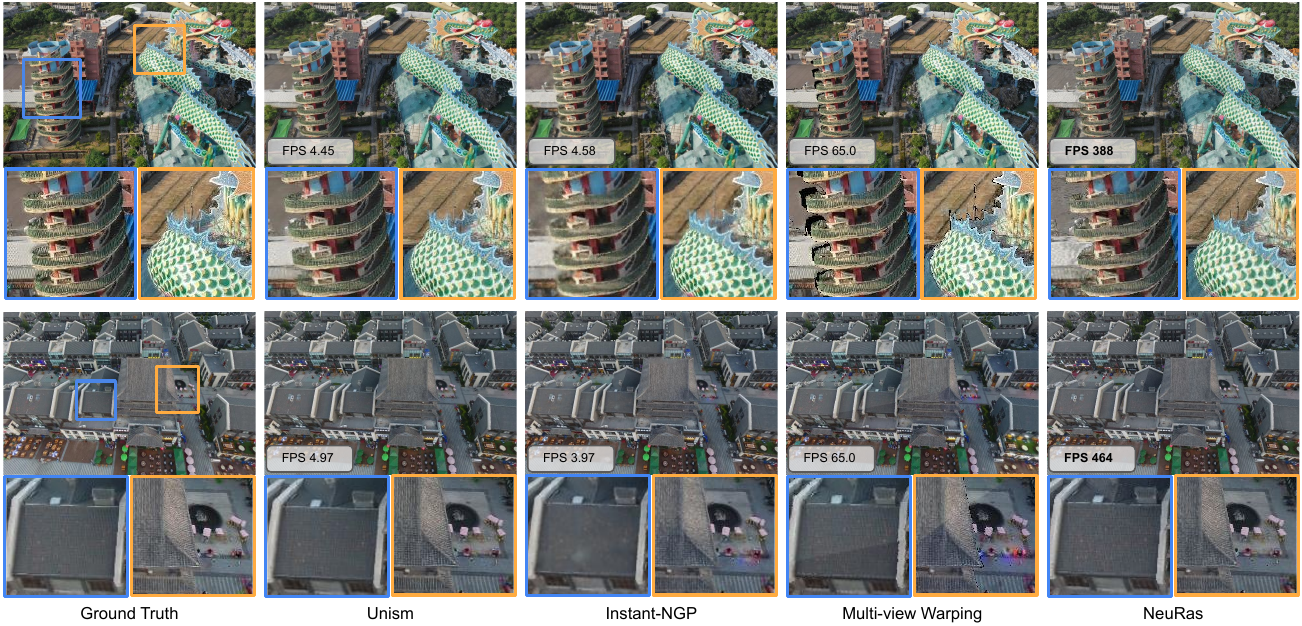}
     \caption{
     \textbf{Qualitative results on drone scenes.} Compared to existing
     novel view synthesis approaches, \methodname produces competitive realism results and achieves real-time rendering ($> 400$ FPS) on drone scenes.
     }
     \label{fig:qual_blend}
\end{figure*}

\subsection{Experimental Setup}

\paragraph{Datasets:}
We conduct experiments primarily on two public datasets with large-scale scenes: PandaSet~\cite{xiao2021pandaset} and BlendedMVS~\cite{yao2019blendedmvs}.
PandaSet is a real-world driving dataset that contains $103$ urban scenes captured in San Francisco, each with a duration of $8$ seconds ($80$ frames, sampled at $10$Hz) and a coverage of around $300\times80 m^2$ meters.
At each frame, a front-facing camera image ($1920 \times 1080$) and $360$ degree point cloud are provided.
We select $9$ scenes on Pandaset that have few dynamic actors for NVS evaluation.
Since
our focus is on
static scenes, the dynamic actor
regions are
masked out during evaluation by projecting the 3D bounding boxes into images.
The BlendedMVS dataset offers a collection of large-scale outdoor scenes captured by a drone and also provides reconstructed meshes generated from a 3D reconstruction pipeline~\cite{altizure}.
We select $5$ large scenes for evaluation which are diverse and range in size from $200\times 200m^2$ to over $500\times 500m^2$.

\vspace{-5pt}

\paragraph{Baselines:}
We compare our approach with two types of novel view synthesis (NVS) methods:
(1) \textit{Implicit-based neural fields}: Instant-NGP~\cite{mueller2022instant}, ENeRF~\cite{lin2022efficient}, UniSim~\cite{unisim}.
Instant-NGP introduces a multi-resolution hashing grid and tiny MLP for fast training and inference.
UniSim and ENeRF leverage the LiDAR or depth information for efficient ray generation (sparse grids) and efficient sampling along each ray.
(2) \textit{Explicit-based approaches}: multi-view warping~\cite{chaurasia2013depth} uses the mesh to project nearby source images to target views, and FVS~\cite{riegler2020fvs} further uses neural networks to blend the source images.

\vspace{-10pt}

\paragraph{Implementation details:}
In our implementation, we utilized a UV feature size of $8192 \times 8192$ with $12$ channels for the neural texture component.
Additionally, we employed $6$ skyboxes placed at a distance
ranging
from $150m$ to $2.4km$ from the inner cuboid center.
The MLP consists of $3$ layers, with each layer comprising of $32$ hidden units.
Both codebooks have a size of $1024$.
We trained the model using the Adam optimizer with a learning rate of $0.01$ for $20K$ iterations.
We use a single machine equipped with an A5000 GPU for all reported runtimes and memory usage benchmarking, including the baselines.
Further implementation details are provided in the supp. materials.

\begin{table}[t]
	\centering
	\resizebox{0.99\linewidth}{!}{
		\begin{tabular}{llllll}
			\toprule
			Methods
			& {PSNR$\uparrow$ } & {SSIM$\uparrow$ } & {LPIPS$\downarrow$ }  & FPS$\uparrow$  \\
			\midrule
			Instant-NGP~\cite{mueller2022instant} &24.40   &0.780    &0.241  &8.7  \\
			UniSim~\cite{unisim}&24.37   &0.792   &0.204    &4.9   \\
			\midrule
			Multi-View Warping~\cite{chaurasia2013depth} & 21.11&0.738&0.248  & 65  \\
			 \rowcolor{grey} Ours&24.19   &0.790    &0.176   & 462  \\
			\bottomrule
	\end{tabular}}
	\vspace{5pt}
	\caption{\textbf{State-of-the-art comparison on BlendedMVS}.
	}
	\vspace{-5pt}
	\label{tab:sota_blendedmvs}
\end{table}

\subsection{Fast Rendering on Large-scale Scenes}

\paragraph{Driving scenes on PandaSet:}
For self-driving, we need to simulate the camera images at significantly different viewpoints (\textit{interpolation} and \textit{extrapolation}).
We evaluate the \textit{interpolation} setting following~\cite{liao2022kitti}: sub-sampling the sensor data by two, training on every other frame and testing on the remaining frames.
We report PSNR, SSIM~\cite{wang2004image}, and LPIPS~\cite{zhang2018unreasonable}.
Moreover, we evaluate a more challenging \textit{extrapolation} setting (\textit{Lane Shift}) following~\cite{unisim} by simulating a new trajectory shifted laterally to the left or right by 2 or 3 meters.
We report FID~\cite{parmar2022aliased} since there is no ground truth for comparison.
We use the neural meshes reconstructed by UniSim~\cite{unisim} for the experiments.

As shown in Fig.~\ref{fig:tradeoff}, our approach strikes the best trade-off between rendering quality and rendering speed.
Table~\ref{tab:sota_panda} shows all the metrics, specifically, our method only sacrifices $0.8$ PSNR compared to the best approach while maintaining $> 100$ FPS ($80 \times$ faster than UniSim, $40 \times$ faster than Instant-NGP).
In contrast, implicit-based neural fields obtain slightly better (UniSim) or worse (InstantNGP, ENeRF) PSNR with much lower inference speed.
Compared to geometry-based approaches such as multi-view warping and FVS, \methodname achieves better realism and faster rendering speed.
Furthermore, the qualitative results presented in Fig.~\ref{fig:qual_panda} demonstrate that our method exhibits comparable or superior visual realism when compared to the baselines.

\paragraph{Drone scenes on BlendedMVS:}
We leverage the provided reconstructed meshes in BlendedMVS as our geometry scaffold and evaluate interpolation setting (\ie, random $50-50$ train-validation split).
As shown in Table~\ref{tab:sota_blendedmvs}, \methodname strikes a good balance between rendering quality and speed.
Fig.~\ref{fig:qual_blend} shows qualitative comparisons with state-of-the-art NVS methods.
Except for view-warping, which has visible artifacts, it is difficult to discern differences between our method and slower neural rendering methods.

\paragraph{Method Ablation:}
We ablate our design choices for \methodname on neural shader and vector quantization (VQ).
As shown in Table~\ref{tab:ablate_mlp}, using an MLP-backed fragment shader can improve realism in generating view-dependent results.
It can also help compensate for artifacts in the geometry.
Please refer to supp. for visual comparison.
In Table~\ref{tab:ablate_VQ}, we show vector quantization significantly reduce the disk storage while maintaining similar realism.
Besides, adding VQ
helps
regularize the neural texture maps thus resulting in better perceptual quality especially in extrapolation results (indicated by smaller FID in the lane-shift setting).

\begin{figure}[]
	\centering
	\includegraphics[width=0.45\textwidth]{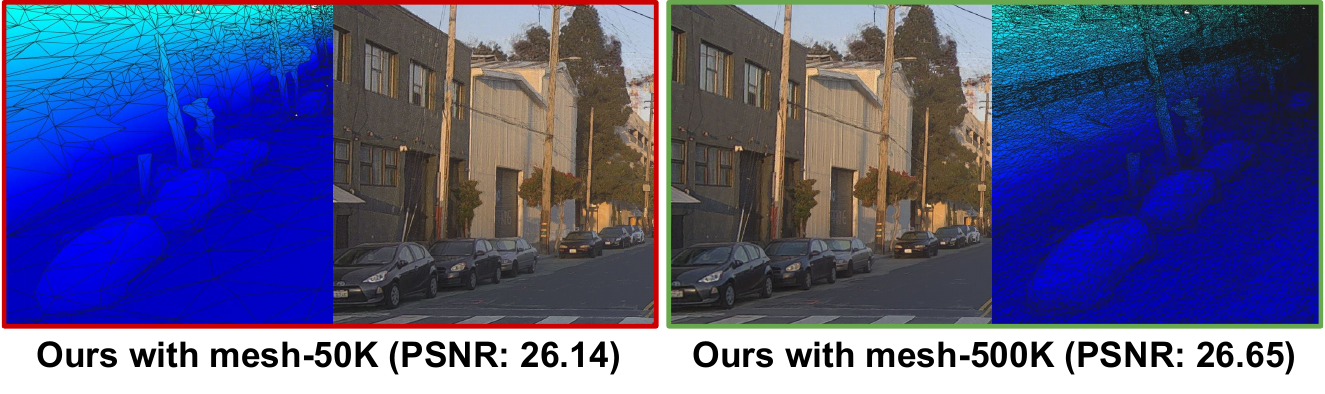}
	\caption{\textbf{Qualitative comparison between using mesh with 50K and 500K vertices}}
	\label{fig:mesh_ablate_qualitative}
\end{figure}

\paragraph{Mesh Ablation:}%
We also demonstrate that \methodname can perform well with coarser low poly-count meshes in Fig~\ref{fig:mesh_ablate_qualitative}.
Given the geometry mesh extracted with UniSim \cite{unisim}, we perform triangle decimation \cite{garland1997surface} to achieve desired vertex counts, and then learn the texture maps with \methodname.
Our approach achieves similar visual quality despite the low resolution mesh, desmonstrating the value of the neural textures.

\begin{table}[t]
	\centering
	\resizebox{0.45\textwidth}{!}{
		\begin{tabular}{lcccc}
			\toprule
			Methods & {PSNR$\uparrow$ } & {SSIM$\uparrow$ } & {LPIPS$\downarrow$}   \\
			\midrule
			No MLP &  24.48 & 0.664 & 0.374  \\
			MLP-shader w/o viewdir  & 25.01  & 0.728  &0.316  \\
			MLP shader & \textbf{25.34} & \textbf{0.738}  & \textbf{0.308} \\
			\bottomrule
		\end{tabular}}
	\vspace{2pt}
	\caption{\textbf{Ablation on MLP shader.} Metrics are reported on the \texttt{log-53} in PandaSet.}
	\label{tab:ablate_mlp}
\end{table}

\begin{table}[t]
	\centering
	\resizebox{\linewidth}{!}{
		\begin{tabular}{lccccc}
			\toprule
			\multirow{2}{*}{Methods} & \multicolumn{2}{c}{Interpolation} &
			\multicolumn{2}{c}{Lane Shift} & \multirow{2}{*}{Storage $\downarrow$} \\
			\cmidrule(r){2-3} \cmidrule(l){4-5}
			& {PSNR$\uparrow$ } & {LPIPS$\downarrow$} &  {FID$\downarrow$@2m} &  {FID$\downarrow$@3m}  & (MB) \\
			\midrule
			w/o VQ &  \textbf{25.46} & 0.317 & 81.5 & 98.3 & 4644 \\
		    w/ VQ  & 25.34 & \textbf{0.308} & \textbf{78.2} & \textbf{95.3} & \textbf{394} \\
			\bottomrule
	\end{tabular}}
	\vspace{2pt}
	\caption{\textbf{Ablation on vector quantization.} Vector quantization improves extrapolation and reduces storage, with minimal impact on realism. Metrics are reported on \texttt{log-53} in PandaSet.}
	\label{tab:ablate_VQ}
\end{table}

\vspace{-12pt}

\paragraph{Comparison against real-time NeRFs:}
When comparing our method with real-time rendering approaches such as SNeRG~\cite{hedman2021baking}, PlenOctrees~\cite{yu2021plenoctrees}, and MobileNeRF~\cite{chen2022mobilenerf}, we encountered difficulties in applying these methods to large scenes and achieving satisfactory results. These challenges included low-resolution representation, out-of-memory errors, limited model capacity, high training cost, and poor geometry.
We therefore compare against these methods on object-level scenes in Table~\ref{tab:nerf_synthetic}.
We use VoxSurf \cite{wu2023voxurf} to extract the geometries for \methodname.
Please see supp. for more details.
While we focus on the real-time rendering for large scenes, our approach has reasonable performance on small-objects from NeRF synthetic dataset~\cite{mildenhall2020nerf}.

\begin{table}[t]
	\setlength{\tabcolsep}{2.5pt}
	\resizebox{\linewidth}{!}{
		\begin{tabular}{lccccccccc}
			\toprule
			\multirow{2}{*}{Methods} &  \multicolumn{3}{c}{Chair} & & \multicolumn{3}{c}{Lego} & \multirow{2}{*}{\ FPS$\uparrow$} \\ \cline{2-4} \cline{6-8}
			& {PSNR$\uparrow$ } & {SSIM$\uparrow$ } & {LPIPS$\downarrow$} & & {PSNR$\uparrow$ } & {SSIM$\uparrow$ } & {LPIPS$\downarrow$}   \\
			\midrule
			NeRF~\cite{mildenhall2020nerf}       & 33.00  & 0.967   & 0.046 &   &  32.54   & 0.961 & 0.050 & 0.02 \\
			Mip-NeRF~\cite{barron2021mip}       & 37.14  & 0.988   & 0.011 &   &  35.74   & 0.984 & 0.013 &  {0.02 }\\ %
			TensoRF~\cite{chen2022tensorf}       &  35.76   & 0.985   & 0.022 &   & 36.46   & 0.983 & 0.018 &   1.5 \\ %
			NSVF ~\cite{liu2020neural}  & 33.19  & 0.968   & 0.043 &   &  32.29   & 0.973  & 0.029  & 0.84 \\ \midrule
			SNeRG~\cite{hedman2021baking}    &  33.24 & 0.975 & 0.025 &   &  33.82 &  0.973 & 0.022 & 176  \\
			PlenOctree~\cite{yu2021plenoctrees}    & 33.19  &  0.970   & 0.039  &  & 32.12 & 0.965 & 0.046  & 270  \\
			MobileNeRF~\cite{chen2022mobilenerf}   & {34.02} & {0.978}  & {0.025} &     &  {34.18}  &  {0.975}  &  {0.025}  &  {720} \\ \midrule
			Ours             & 33.15  & 0.975  & 0.036  &      &  30.66 &  0.951 & 0.061 &  461 \\
			\bottomrule
	\end{tabular}}
	\vspace{2pt}
	\caption{\textbf{Comparison with real-time NeRFs on synthetic dataset.} Despite being designed for large-scale scenes, our method still achieves comparable performance when rendering small objects.}
	\label{tab:nerf_synthetic}
\end{table}

\subsection{Speeding-up NeRFs}
\vspace{5pt}
\begin{figure*}[t]
	\begin{center}
	\includegraphics[width=1.0\textwidth]{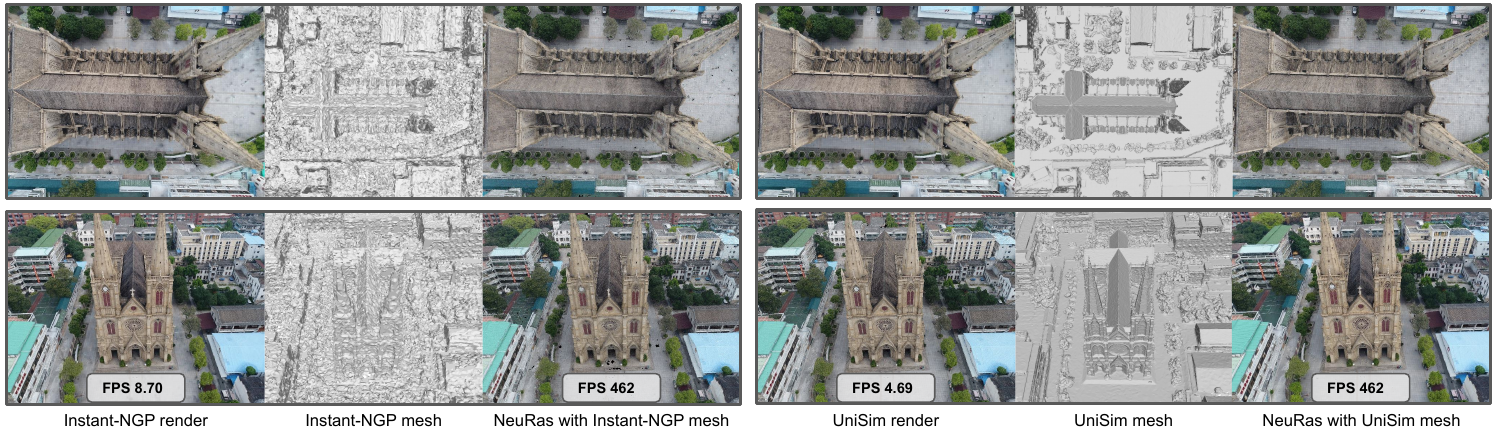}
\end{center}
\caption{
\textbf{Speed up neural radiance fields with \methodname.}
We use Marching Cubes~\cite{lorensen1987marching} to extract the geometry from the trained radiance fields model and then adopt NeuRas for real-time rendering.
For each example, we show radiance field rendering in the left, extracted mesh in the middle, and \methodname rendering in the right.
\methodname can significantly speed up the rendering speed while maintaining a similar rendering photorealism even with poor geometry scaffold (\eg Instant-NGP~\cite{mueller2022instant}).
}
\label{fig:speed_nerf}
\end{figure*}

\label{sec:analysis}
We highlight that \methodname can speed up popular NeRF approaches.
We consider two representative approaches: Instant-NGP (camera supervision, density geometry) and UniSim (camera + depth supervision, SDF geometry).
We use marching cubes~\cite{lorensen1987marching} to extract the geometry for a selected scene in BlendedMVS and then adopt \methodname for real-time rendering.
As shown in Table~\ref{tab:speedup_nerf} and Fig~\ref{fig:speed_nerf}, our approach dramatically speeds up rendering performance while maintaining reasonable photorealism.
Please see supp. materials for visual comparison and more analysis.

\begin{figure*}[]
	\begin{center}
	\includegraphics[width=0.99\textwidth]{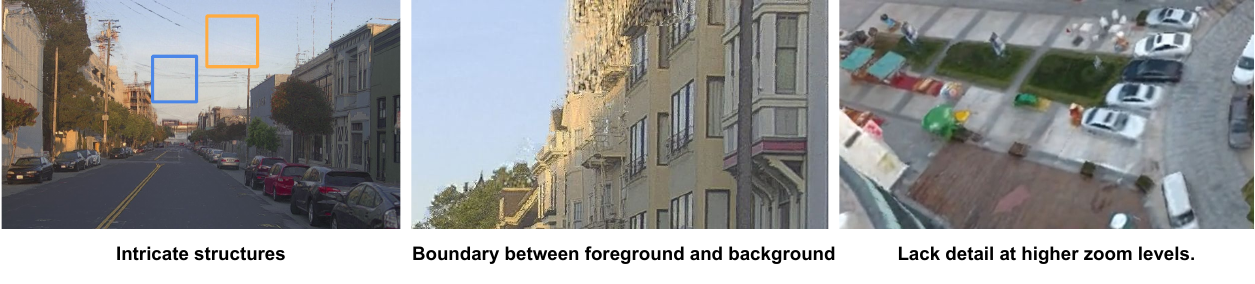}
\end{center}
\caption{
\textbf{Limitations of \methodname.}
From left to right: Difficulties in modelling intricate structures such as power lines in the sky, artifacts at the boundary of foreground mesh and skybox due to missing geometry at the top of building, and lack of details at high zoom levels. %
}
\label{fig:supp_limitations}
\end{figure*}

\begin{table}[t]
	\centering
	\resizebox{0.45\textwidth}{!}{
		\begin{tabular}{lcccr}
			\toprule
			Methods
			& {PSNR$\uparrow$ } & {SSIM$\uparrow$ } & {LPIPS$\downarrow$}  & FPS$\uparrow$ \\
			\midrule
			InstantNGP~\cite{mueller2022instant}                    & 23.55                 & 0.75                 & 0.23                     &   8.2  \\
			Ours + InstantNGP             & 20.26                 & 0.52                 & 0.31                   &    455 \\
			\midrule
			UniSim~\cite{unisim}                           & 23.30                & 0.76                 & 0.21                  &  4.7   \\
			Ours + UniSim                 & 21.19                & 0.65                  & 0.23                   &   336  \\
			\bottomrule
			\end{tabular}}
	\vspace{1pt}
	\caption{\textbf{Speed up neural radiance fields.} NeuRas is able to speed up popular NeRFs even if the geometry is not high-quality. The metrics are reported on \texttt{Church} in BlendedMVS.}
	\vspace{-5pt}
	\label{tab:speedup_nerf}
	\end{table}

\subsection{Limitations}

Our method has several limitations, including the use of opaque meshes, which poses challenges in accurately modeling semi-transparent components such as fog and water.
Additionally, while our neural shader design improves the robustness of the model
{\wrt}
geometry quality, rendering performance may still suffer when dealing with meshes with severe artifacts
 in Fig.~\ref{fig:supp_limitations}.
Our method cannot
fix severe mesh artifacts and incorrect boundaries, which
requires a fully differentiable rasterization pipeline.
Additionally, our approach has difficulty rendering completely
unseen regions that are far from the training views.
Scene completion may help address this.
Moreover, the current implementation uses only one UV
texture level,
which can cause aliasing or blurriness when
scaling extensively, such as when moving very close to or
far away from an object. Using a multi-level texture representation such as mipmaps \cite{williams1983pyramidal} could mitigate these issues.
More analysis is available in the supp. materials.

%% file: tex/5-conclusion.tex
\vspace{3pt}
\section{Conclusion}

In this paper, we present \methodname, a novel approach for realistic real-time novel view synthesis of large scenes.
Our approach combines the strengths of
neural rendering and traditional %
graphics %
to achieve the best trade-off between realism and efficiency.
\methodname utilizes a scaffold mesh as input and incorporates a neural texture field to model
view-dependent effects, which can then be exported and rendered in real-time with standard rasterization engines.
\methodname can render urban driving scenes at $1920 \times 1080$ resolution
at over 100 FPS while delivering comparable realism to existing neural rendering approaches.
We hope \methodname can open up possibilities for scalable and immersive experiences for self-driving simulation and VR applications.

\paragraph{Acknowledgements:} We thank Ioan-Andrei Barsan for
profound discussion and constructive feedback. We thank
the Waabi team for their valuable assistance and support.